%% file: IEEE-conference-template-062824.tex
\documentclass[10pt,conference]{IEEEtran}
\IEEEoverridecommandlockouts

\usepackage{cite}
\usepackage{amsmath,amssymb,amsfonts}
\usepackage{graphicx}
\usepackage{textcomp}
\usepackage{xcolor}

\usepackage{booktabs}
\usepackage{multirow}
\usepackage{graphicx}
\usepackage[table]{xcolor}
\usepackage{arydshln}

\usepackage{algorithm}
\usepackage{algpseudocode}
\usepackage{algorithmicx}
\usepackage{tikz}
\usepackage{amsmath}
\usepackage[hidelinks]{hyperref}

\def\BibTeX{{\rm B\kern-.05em{\sc i\kern-.025em b}\kern-.08em
    T\kern-.1667em\lower.7ex\hbox{E}\kern-.125emX}}
\begin{document}

\title{IE as Cache: Information Extraction Enhanced Agentic Reasoning
}



\author{
 \textbf{Hang Lv\textsuperscript{1}},
 \textbf{Sheng Liang\textsuperscript{2}}, 
 \textbf{Hongchao Gu\textsuperscript{1}},
 \textbf{Wei Guo\textsuperscript{2}},
 \textbf{Defu Lian\textsuperscript{1}},\\
 \textbf{Yong Liu\textsuperscript{2}},
 \textbf{Hao Wang\textsuperscript{1}},
 \textbf{Enhong Chen\textsuperscript{1}}
 \\
 \textsuperscript{1}University of Science and Technology of China, 
 \textsuperscript{2}Huawei Technologies Co., Ltd.
}

\maketitle

\begin{abstract}

Information Extraction aims to distill structured, decision-relevant information from unstructured text, serving as a foundation for downstream understanding and reasoning. However, it is traditionally treated merely as a terminal objective: once extracted, the resulting structure is often consumed in isolation rather than maintained and reused during multi-step inference. Moving beyond this, we propose \textit{IE-as-Cache}, a framework that repurposes IE as a cognitive cache to enhance agentic reasoning. Drawing inspiration from hierarchical computer memory, our approach combines query-driven extraction with cache-aware reasoning to dynamically maintain compact intermediate information and filter noise. Experiments on challenging benchmarks across diverse LLMs demonstrate significant improvements in reasoning accuracy, indicating that IE can be effectively repurposed as a reusable cognitive resource and offering a promising direction for future research on downstream uses of IE.


\end{abstract}


\input{sections/sec01-introduction}

\input{sections/sec02-related_work}

\input{sections/sec03-method}

\input{sections/sec04-experiment}
\input{sections/sec05-conclusion}
\bibliographystyle{IEEEtran}
\bibliography{ref}
\newpage
\input{sections/secapp}

\end{document}

%% file: sections/sec01-introduction.tex
\section{Introduction}

In the landscape of web search and natural language understanding, Information Extraction (IE) stands as a pivotal technique for parsing the vast ocean of unstructured digital content into structured knowledge~\cite{ie1,ie2,ie3} such as entities~\cite{ner1,ner2}, relations~\cite{re1,re2}, and events~\cite{ee1,ee2}. Despite its practical importance, contemporary research predominantly treats IE as an end task—optimizing extraction quality as the final objective—rather than exploring IE as a reusable infrastructure that can be continuously consumed and updated by downstream cognitive processes~\cite{llmiesurvey}. In this work, we posit that strategically extracted and maintained information can directly scaffold large language model (LLM) reasoning, especially as LLMs increasingly operate in an agentic manner where they must iteratively read, decide, and act over complex inputs \cite{agenticreasoning}.

This perspective is crucial for processing noise-rich, long-form content~\cite{tay2020longrangearenabenchmark}, where LLMs struggle with irrelevant distractors~\cite{distraction} and information decay in middle contexts~\cite{lostinthemiddle}. In such settings, simply providing raw context is often insufficient; instead, models benefit from a compact intermediate representation that preserves salient evidence while suppressing noise~\cite{icl}. Addressing these inefficiencies, we draw inspiration from hierarchical computer memory~\cite{memgpt} to introduce the cognitive analogy in Figure~\ref{fig:1}. Just as CPU caches maintain critical data for rapid retrieval, we hypothesize that IE can function as a dynamic memory layer that filters noise and preserves semantically salient content for reasoning.

\begin{figure}
    \centering
    \includegraphics[width=\linewidth]{figures/Fig1_Version_2.png}
    \caption{Cognitive analogy between hierarchical computer memory and text reasoning. Information Extraction functions as a bridge between raw text and the reasoning agent, mirroring the role of a hardware cache.}
    \label{fig:1}
\end{figure}

This analogy leads to our central research question: \textbf{\textit{Can IE serve as a cognitive cache to scaffold complex reasoning processes?}} To explore this hypothesis, we propose the \textit{IE-as-Cache} framework, which consists of two coordinated components: \textit{Query-Driven Information Extraction} and \textit{Cache-aware Agentic Reasoning}. Given a user request, the framework first extracts query-relevant content to initialize a compact, structured cache. This cache serves as the primary working memory, enabling the agent to operate on a noise-reduced representation rather than repeatedly rescanning the full text. During the subsequent reasoning process, the model dynamically accesses raw data on-demand to update the cache with new evidence, while strictly controlling redundancy. Our experiments demonstrate the framework's effectiveness on challenging benchmarks, highlighting the potential of IE to significantly improve LLM reasoning as a reusable cognitive resource. 

To sum up, our contributions are threefold:

\begin{itemize}

\item  We propose a new viewpoint that repurposes information extraction from an endpoint task to an explicit cognitive cache that actively scaffolds agentic reasoning.

\item we develop \textit{IE-as-Cache}, a noval framework that integrates query-driven extraction with cache-aware reasoning and dynamic cache updates for noise-robust inference.

\item we conduct extensive experiments across diverse LLMs and three representative task families (question answering, planning, and query-focused summarization), demonstrating consistent improvements in reasoning accuracy.

\end{itemize}

\begin{figure*}
    \centering
    \includegraphics[width=\linewidth]{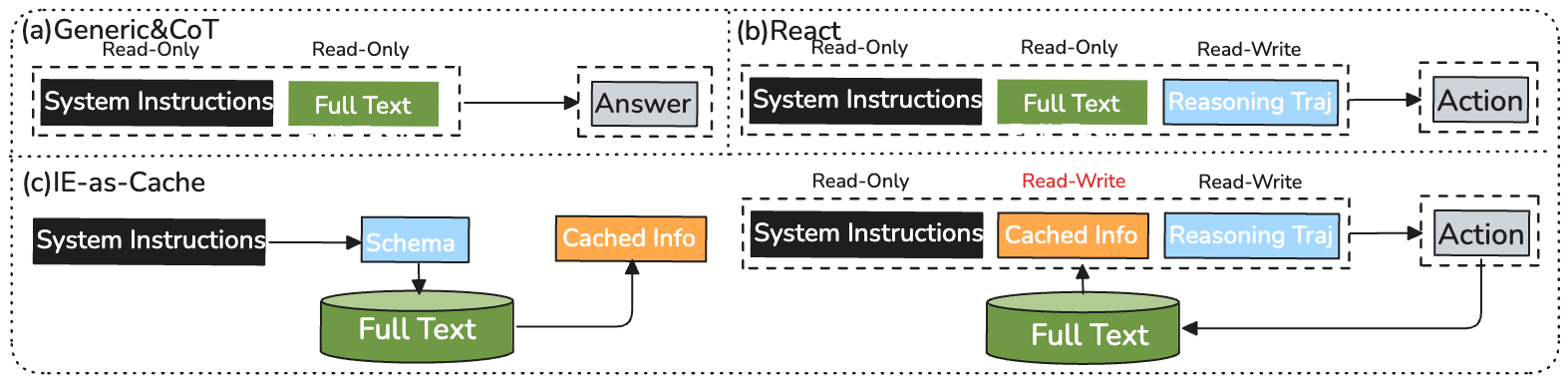}
    \caption{Comparison of context mechanisms. Unlike (a) and (b) which treat raw text as a static read-only block, (c) IE-as-Cache introduces a dynamic Read-Write Cache. This mechanism filters noise by keeping the full text external and only maintaining semantically critical information in the active memory.}
    \label{fig:2}
\end{figure*}

%% file: sections/sec02-related_work.tex
\section{Related Work}

\paragraph{\textbf{Information Extraction as Foundational Infrastructure}} The evolution of information extraction (IE) has progressed from task-specific solutions to more unified frameworks. Early work laid the foundation with core tasks like Named Entity Recognition \cite{ner1}, Relation Extraction \cite{re1}, and Event Extraction \cite{ee1}, each requiring specialized architectures and limited to specific domains. The paradigm shift came with Unified Information Extraction \cite{unifiedie}, which demonstrated that schema-guided prompting could harmonize these diverse extraction tasks into structured templates, substantially reducing task fragmentation. However, despite this unification, these approaches have predominantly treated IE as an \textit{end task}, emphasizing extraction quality as the final deliverable, which limits its use as a \textit{reusable cognitive resource} for ongoing reasoning and multi-step cognitive processes.

\paragraph{\textbf{Operationalizing IE for Knowledge Systems}} 
While traditional research has used IE to construct knowledge graphs and extract structured information for specific tasks~\cite{application-knowledgegraph, kgsurvey}, its role in supporting dynamic, multi-step reasoning in large language models (LLMs) remains underexplored~\cite{reasoning}. Some efforts leverage IE outputs for domain-specific applications, such as scientific knowledge discovery and fact-based search~\cite{application-knowledgediscovery, zhang-etal-2022-deepke}, organizing extracted facts as relatively stable artifacts for downstream querying. However, these pipelines typically treat IE as a one-time process, and the extracted structures are rarely maintained or updated for iterative reasoning, limiting their flexibility and scalability for general-purpose LLM reasoning~\cite{iterative}.

\paragraph{\textbf{Structured Reasoning Architectures}} Recent advances in prompting highlight the critical role of structured information in enhancing LLM reasoning capabilities. Studies show that \textit{template structures} can significantly affect reasoning fidelity~\cite{formatimpact}. The Chain-of-Table paradigm~\cite{chainoftable} leverages tabular representations for numerical reasoning, while~\cite{tableasthought} formalizes \textit{table-structured thought} for multi-task generalization. However, these methods largely emphasize \textit{output formatting} and intermediate traces, leaving a gap in optimizing the \textit{input}. Our work addresses this gap by utilizing IE as a cognitive cache to maintain a compact, noise-resistant representation that can be retrieved and updated during multi-step reasoning.

%% file: sections/sec03-method.tex
\section{IE as Cache}

In this section, we introduce \textit{IE-as-Cache}, a framework that repurposes Information Extraction (IE) as a cognitive cache for agentic reasoning. Given a user query $Q$ and a raw text source $T$, the goal is to produce a final answer $A$ while reducing the model's reliance on repeatedly reading long, noise-rich inputs. The key idea is to treat the raw text as external storage and maintain a compact, read-write cache $C$ inside the agent's working context. This framework consists of two coordinated components: \textit{Query-Driven Information Extraction} for initializing the cache, and \textit{Cache-Aware Agentic Reasoning} for using and updating it during multi-step inference. Algorithm~\ref{alg:iecache} summarizes the overall procedure.

\subsection{Query-Driven Information Extraction}
\label{sec3.1}
Effective cache initialization requires the extraction targets to adapt to the user's intent. Conventional IE pipelines typically rely on fixed schemas (e.g., predefined entity/relation sets). Such fixed schemas are often mismatched with open-domain reasoning, where the appropriate extraction granularity depends on the semantics of the query~\cite{ODIE,ODIE2}. We therefore consider two candidates for query-conditioned extraction.

\noindent\textbf{Paradigm 1: Monolithic Extraction}
\begin{equation}
E = \mathcal{F}_{\text{LLM}}(Q, T),
\end{equation}
where an LLM directly maps the query $Q$ and raw text $T$ to extractions $E$. While simple, this formulation entangles \emph{schema identification} (what to extract) with \emph{content extraction} (which facts to extract), often leading to noisy, inconsistent, or overly specific structures~\cite{liang2025adaptiveschemaawareeventextraction}.

\noindent\textbf{Paradigm 2: Schema-Decoupled Extraction}
\begin{align}
S &= \mathcal{F}_{\text{LLM}}^{\text{schema}}(Q), \\
E &= \mathcal{F}_{\text{LLM}}^{\text{extract}}(Q, S, T),
\end{align}
where we first generate a query-conditioned extraction schema $S$, and then perform schema-guided extraction to obtain $E$. Intuitively, $S$ specifies the extraction structure aligned with the query (e.g., slot names or table columns), and the schema-guided extractor focuses the model on extracting query-relevant content from $T$ under this structure. In practice, we represent $E$ as a compact set of structured entries (e.g., a table-like record set), which is more suitable as an intermediate cache than free-form text.

Our preliminary experiments showed that this two-stage approach significantly outperforms monolithic extraction, so we adopt it for initializing the cache. We then form the agent's initial prompt by combining the user query $Q$ with the extracted structure $E$, which serves as the initial cache for subsequent reasoning.

\subsection{Cache-Aware Agentic Reasoning}
\label{sec3.2}
While \S\ref{sec3.1} provides an initial cache, the core of \textit{IE-as-Cache} is how an agent reasons with $C$ and maintains it dynamically. As illustrated in Figure~\ref{fig:2}, we shift from a static context paradigm to a cache-centric design.

\subsubsection{\textbf{Architectural Comparison: Standard agent vs. IE-as-Cache}}
The fundamental distinction between our approach and standard reasoning agents (e.g., ReAct \cite{react}) lies in how the context window is allocated.

\noindent \textbf{Standard ReAct Architecture (Figure~\ref{fig:2}b):}
A standard agent places the raw text $T$ into the context window as a read-only block and appends observations and reasoning traces during interaction. For long-form inputs, this design introduces substantial redundancy and distractors, and the agent repeatedly scans the raw text at each step. As a result, the working context is dominated by uncompressed input, and the dynamic component is primarily the reasoning trajectory.

\noindent \textbf{IE-as-Cache Architecture (Figure~\ref{fig:2}c):}
We instead treat the full text $T$ as external storage and keep a compact, structured cache $C$ as the primary read-write object in working memory. The agent reasons over $C$, and only accesses $T$ on demand. Importantly, the cache is not a static summary: it is explicitly maintained and updated as reasoning progresses, preserving only semantically salient, query-relevant information in the agent's immediate focus.

\subsubsection{\textbf{Reason--Act Interface}}
At each step, the agent produces a reasoning trace and, when needed, triggers an action:
\begin{equation}
(r, a) \leftarrow \mathcal{F}_{\text{LLM}}^{\text{reason}}(Q, C).
\end{equation}
We implement \texttt{seek\_information} as a lightweight action within a standard agentic loop, enabling step-specific evidence acquisition and cache maintenance with minimal engineering changes. The overall workflow remains compatible with existing ReAct-style agents (Algorithm~\ref{alg:iecache}).
\input{tables/algo}

\subsubsection{\textbf{On-Demand Extraction as Memory Access}}
When the agent issues $\texttt{seek\_information}(q_t)$, it does not retrieve and append raw snippets. Instead, we reuse the schema-decoupled extractor to perform \emph{on-demand extraction} over the raw text:
\begin{equation}
E' \leftarrow \mathcal{F}_{\text{LLM}}^{\text{extract}}(Q, S, T, q_t),
\end{equation}
where $q_t$ provides a step-specific focus indicating what information should be extracted to resolve the current uncertainty. This design aligns with the cache analogy: the agent accesses external storage and reads back only structured, query-relevant content rather than unstructured observations.

\subsubsection{\textbf{Cache Update: Action $\rightarrow$ Cache Maintenance}}
After obtaining the newly extracted $E'$, we update the cache via
\begin{equation}
C \leftarrow \mathcal{F}_{\text{LLM}}^{\text{update}}(Q, S, C, E').
\end{equation}
The update operator synthesizes newly extracted content with existing cache entries. To keep this step lightweight, we perform cache maintenance in a single LLM call: the model (i) merges compatible information into existing entries when appropriate, (ii) removes redundant or off-target content, and (iii) prunes low-utility details to keep the cache compact. This maintenance step is the key difference from standard agents: it replaces ``Action $\rightarrow$ Observation'' with an explicit ``Action $\rightarrow$ Cache Update'' cycle, so the agent reasons over a compact, noise-resistant representation throughout multi-step inference.


\subsubsection{\textbf{Self-Check \& Fallback}}
Optionally, the agent performs a lightweight self-check to revise the cache or refine the next information request, without changing the main loop (Algorithm~\ref{alg:iecache}). The reasoning process terminates when the agent outputs $\texttt{final}(A)$ or reaches a maximum of $H$ steps; if the step limit is reached, we generate the answer from the current cache as a fallback:
\begin{equation}
A \leftarrow \mathcal{F}_{\text{LLM}}^{\text{answer}}(Q, C).
\end{equation}

Overall, \textit{IE-as-Cache} turns information extraction into a lightweight, dynamically maintained cache that keeps only query-relevant structure in the working context and updates it on demand, helping agents reason over long, noise-rich text without repeatedly re-reading the full input and reducing distraction during multi-step inference.

%% file: tables/algo.tex
\begin{algorithm}[t]
\caption{IE-as-Cache Framework}
\label{alg:iecache}
\begin{algorithmic}[1]
    \Require Query $Q$, raw text $T$, max steps $H$
    \Ensure Answer $A$

    \State \textbf{Query-Driven Extraction (Initialize Cache)}
    \State $S \leftarrow \mathcal{F}^{\text{schema}}_{\text{LLM}}(Q)$ \Comment{query-conditioned schema}
    \State $E \leftarrow \mathcal{F}^{\text{extract}}_{\text{LLM}}(Q, S, T)$ \Comment{schema-guided extraction}
    \State $C \leftarrow \texttt{InitCache}(Q, E)$

    \State \textbf{Cache-Aware Agentic Reasoning}
    \State $t \leftarrow 0$
    \While{$t < H$}
        \State $(r, a) \leftarrow \mathcal{F}^{\text{reason}}_{\text{LLM}}(Q, C)$
        \If{$a = \texttt{final}(A)$}
            \State \Return $A$
        \EndIf
        \If{$a = \texttt{seek\_information}(q_t)$}
            \State $E' \leftarrow \mathcal{F}^{\text{extract}}_{\text{LLM}}(Q, S, T, q_t)$ \Comment{on-demand extraction}
            \State $C \leftarrow \mathcal{F}^{\text{update}}_{\text{LLM}}(Q, S, C, E')$ \Comment{merge/prune}
            \If{\texttt{NeedCheck}$(r, C)$} \Comment{optional}
                \State $C \leftarrow \mathcal{F}^{\text{check}}_{\text{LLM}}(Q, S, C, r)$ \Comment{self-check}
            \EndIf
        \EndIf
        \State $t \leftarrow t + 1$
    \EndWhile

    \State $A \leftarrow \mathcal{F}^{\text{answer}}_{\text{LLM}}(Q, C)$ \Comment{fallback at step limit}
    \State \Return $A$
\end{algorithmic}
\end{algorithm}

%% file: sections/sec04-experiment.tex
\section{Experiment}
\subsection{Experiment Setup}
\paragraph{\textbf{Tasks}} 

To rigorously evaluate generalizability, we focused on benchmarks involving lengthy, unstructured raw text that demand complex aggregative reasoning. Our selection criteria strictly target scenarios where information must be actively synthesized from noise-rich contexts, moving beyond the simple span retrieval or pre-structured inputs typical of standard benchmarks. Based on this, we employed:
\begin{itemize}
    \item \textbf{Logical QA (TACT~\cite{tact}):} A dataset requiring logical deduction over long-form text. It serves as a testbed for the agent's ability to aggregate scattered evidence and perform multi-hop reasoning without pre-defined schemas or tables.
    
    \item \textbf{Agentic Planning (Calendar Scheduling~\cite{naturalplan}):} A complex constraint satisfaction task where the agent must manage and resolve conflicting schedules derived purely from raw natural language descriptions, simulating personal assistant scenarios.
    
    \item \textbf{Query-Focused Summarization (QMSUM~\cite{qmsum}):} A dataset for summarizing specific spans from extensive multi-turn meeting transcripts based on user queries. This tests the framework's capacity to distill and synthesize key information from high-noise dialogue environments.
\end{itemize}

This diverse suite allows us to verify our method's effectiveness across \textbf{Question Answering}, \textbf{Planning}, and \textbf{Summarization} tasks within noise-rich contexts.



\paragraph{\textbf{Models}} 
We verify our framework across a diverse spectrum of model families and parameter scales. Our selection includes two state-of-the-art proprietary models, \texttt{GPT-4o} and \texttt{GPT-4o-mini} , alongside three open-source models catering to different computational constraints: the \texttt{LLaMA3.1-8B-Instruct} ~\cite{llama3} and the \texttt{Qwen-3} series, ranging from the mid-sized \texttt{8B} to the lightweight \texttt{0.6B} ~\cite{qwen3}.



\paragraph{\textbf{Evaluation Metrics}}
We follow the standard evaluation protocols of each benchmark. For TACT and Calendar Scheduling, we report Exact Match (EM), which directly measures whether the produced answer satisfies the target decision/constraints. For QMSUM, we report \texttt{ROUGE-1} \cite{lin2004rouge}, reflecting query-relevant content coverage in the generated summary. In addition to end-task performance, we further analyze the intermediate extraction quality on TACT using ROUGE and Sentence-T5 similarity~\cite{sentencet5} against the provided ground-truth references (Table~\ref{table:extraction_analysis}), enabling a direct connection between cache quality and final reasoning accuracy.

\paragraph{\textbf{Baselines}} 
To comprehensively evaluate the IE-as-Cache framework, we benchmark it against four established paradigms:

\begin{itemize}
    \item \textbf{Generic (Standard Prompting):} The model receives the instruction and full raw context, generating the answer directly without intermediate steps.
    
    \item \textbf{Chain of Thought (CoT) \cite{chainofthought}:} A reasoning-enhanced baseline that encourages the model to generate explicit reasoning traces over the raw text ("Let's think step by step") before deriving the final answer.
    
    \item \textbf{ReAct \cite{react}:} An agentic baseline where the model interleaves reasoning traces with action execution. To ensure a fair comparison, we designed task-specific action spaces tailored to the unique characteristics of each dataset to facilitate effective interaction with the environment.
    
    \item \textbf{IE-as-Tool \cite{tact}:} A specialized pipeline that first converts text into a structured table and then executes Pandas queries. \textit{Note:} Due to its strict dependency on tabularizable logic, this baseline is exclusively applied to the TACT dataset, as it is incompatible with the unstructured, dynamic nature of Calendar Scheduling and QMSUM.
\end{itemize}

\paragraph{\textbf{Experiment Setup}}
Unless otherwise specified, we use greedy decoding with temperature set to $0$ for all models. For proprietary models (e.g., \texttt{GPT-4o} and \texttt{GPT-4o-mini}), we run each setting three times and report the average performance to account for residual nondeterminism in serving. For open-source models, we fix the random seed and use the same inference configuration across all methods.

\subsection{Main Results}
\label{sec:main_results}
\input{tables/main-tact}
Table \ref{tab:main_results_full_rotated} presents the comprehensive performance comparison across three diverse datasets. Overall, the proposed \textit{IE-as-Cache} framework achieves state-of-the-art performance in most experimental settings, consistently outperforming both standard prompting baselines (Generic, CoT) and agentic baselines (ReAct, IE-as-Tool) across model scales ranging from 0.6B to huge GPT-4o.

\paragraph{\textbf{Overall Trends}}
Across all three task families, \textit{IE-as-Cache} shows the most consistent gains when inputs are long and noisy, and when reasoning requires aggregating multiple cues. These improvements are particularly significant for smaller models, indicating that an explicit cache helps mitigate limited robustness in long-context reasoning by concentrating key information. We also observe that purely agentic interaction (ReAct) is not always beneficial: frequent context switching, driven by raw text and incremental observations, can introduce noise, whereas a compact cache offers a more stable representation for multi-step reasoning.

\paragraph{\textbf{Performance on Logical QA}}
On the TACT dataset, which features long contexts filled with irrelevant distractors, our framework demonstrates the most significant improvements. 
With the strong \texttt{GPT-4o} backbone, \textit{IE-as-Cache} achieves an EM score of \textbf{71.77}, surpassing the specialized baseline \texttt{IE-as-Tool} (61.29) by a substantial margin of over 10 points. 
Notably, this advantage is even more pronounced in smaller open-source models. For instance, on \texttt{Llama3.1-8B}, our method improves performance from 14.52 (Generic) and 27.42 (ReAct) to \textbf{38.71}. 
This indicates that the dynamic cache mechanism effectively filters out noise that typically confuses LLMs in long-context logical reasoning, a capability that rigid pipeline approaches fail to match.

\paragraph{\textbf{Performance on Agentic Planning}}
For the Calendar Scheduling task, which requires strict constraint satisfaction, \textit{IE-as-Cache} continues to lead, particularly with capable proprietary models. 
Using \texttt{GPT-4o}, our method achieves \textbf{65.20} EM, significantly outperforming ReAct (57.00). This suggests that maintaining a high-density cache of "scheduled events" helps the agent better detect conflicts than simply reading raw descriptions.
We observe that smaller models (e.g., \texttt{Qwen3-0.6B}, \texttt{Llama3.1-8B}) struggle significantly with this task across all methods due to the inherent difficulty of reasoning with multiple constraints. However, even in these low-resource scenarios, our framework generally maintains a performance edge (e.g., 17.90 vs 13.50 for \texttt{Qwen3-8B}).
\textit{Note:} One exception is observed with \texttt{Llama-3.1}, where CoT slightly outperforms our method. We attribute this to the model's specific sensitivity to the prompt structure in zero-shot extraction scenarios.

\paragraph{\textbf{Performance on Summarization}}
In the query-focused summarization task involving multi-turn meeting transcripts, \textit{IE-as-Cache} proves to be the most robust approach. 
Interestingly, we observed that the standard \texttt{ReAct} agent often underperforms compared to simple \texttt{Generic} or \texttt{CoT} baselines on this dataset (e.g., ReAct 29.38 vs Generic 32.03 on \texttt{GPT-4o}). This is likely because ReAct's frequent context switching fragments the narrative flow of the meeting.
In contrast, our framework consistently improves upon the baselines across all models (e.g., reaching \textbf{35.21} with \texttt{GPT-4o} and \textbf{33.20} with \texttt{Qwen3-8B}). This confirms that our "Action $\rightarrow$ Extraction $\rightarrow$ Cache Update" cycle successfully distills scattered dialogue turns into a coherent representation without losing the global context required for summarization.

\input{tables/extraction}

\subsection{Ablation and Further Analysis}
To identify the sources of our performance gains, we conduct an ablation study on the cache update mechanism and a deeper analysis of the intermediate extraction quality.

\paragraph{\textbf{Impact of Dynamic Cache Update}}
We first examine the necessity of the \textit{Write-Through Cache Update} mechanism. As shown in the ``\textit{w/o update}'' row of Table \ref{tab:main_results_full_rotated}, removing the agent's ability to dynamically refresh the cache results in a consistent performance drop across all models. 
For instance, on TACT with GPT-4o, the EM score declines from \textbf{71.77} to \textbf{65.32}. This degradation highlights that static extraction is insufficient for multi-hop reasoning. Complex tasks require an iterative process where the agent updates its information state based on intermediate reasoning results, confirming that our dynamic maintenance of the cognitive cache is essential for handling evolving contexts.

\paragraph{\textbf{Correlation between Extraction and Reasoning}}
To further understand \textit{why} IE-as-Cache outperforms the baselines, we analyze the quality of the intermediate extracted content on the TACT dataset (Table \ref{table:extraction_analysis}). 
We benchmark our extracted information against the ground-truth tabular references provided in TACT, utilizing ROUGE scores to measure lexical overlap and Sentence-T5-Large to quantify semantic similarity. 

We observe a positive correlation between extraction quality and final reasoning accuracy. \textit{IE-as-Cache} achieves higher semantic similarity (87.11) compared to \textit{IE-as-Tool} (84.90), which directly translates to a +10.48\% improvement in final EM. This confirms that preserving semantically critical information is a prerequisite for accurate downstream reasoning.

\paragraph{\textbf{Efficiency of Dynamic Schema}}
We also introduced an oracle setting, ``\textit{w/ gold schema}'', where extraction is guided by ground-truth schemas. 
Crucially, the results reveal a diminishing return: while the oracle setting yields significantly higher lexical overlap (ROUGE-1: 50.35 vs 31.04), the gain in final reasoning accuracy is remarkably marginal (EM: 73.39 vs 71.77). 
This suggests that our \textit{IE-as-Cache} framework is highly efficient. Even without the perfect lexical matching of the gold schema, our query-driven dynamic extraction successfully captures the vast majority of \textit{reasoning-critical} information. It constructs a ``sufficiently good'' cognitive cache that supports near-optimal reasoning performance, demonstrating the robustness of our generated extraction plans.

%% file: tables/main-tact.tex
\begin{table}[t]
\centering

\fontsize{7pt}{7pt}\selectfont 
\renewcommand{\arraystretch}{0.85} 
\setlength{\tabcolsep}{1.5pt} 
\setlength{\aboverulesep}{1.5pt}
\setlength{\belowrulesep}{2pt}

\caption{Main experimental results across three datasets. We report Exact Match score for TACT and Calendar, and ROUGE-1 score for QMSUM. Best results are marked in \textbf{bold}.}
\label{tab:main_results_full_rotated}

\resizebox{\linewidth}{!}{%
\begin{tabular}{clccccc}
\toprule
\textbf{Data} & \textbf{Method} & \textbf{GPT-4o} & \textbf{4o-mini} & \textbf{Llama8B} & \textbf{Qwen8B} & \textbf{Qwen.6B} \\
\midrule

\multirow{6}{*}{\rotatebox[origin=c]{90}{\textbf{TACT}}} 
 & Generic & 42.74 & 32.26 & 14.52 & 25.81 & 6.45 \\
 & CoT & 52.42 & 37.90 & 25.00 & 45.97 & 15.32 \\
 & React & 57.26 & 52.42 & 27.42 & 57.26 & 12.90 \\
 & IE as Tool & 61.29 & 45.16 & 31.45 & 26.29 & 12.10 \\
 \cdashline{2-7} 
 \noalign{\vskip 1.5pt} 
 & \textbf{IE as Cache} & \textbf{71.77} & \textbf{58.87} & \textbf{38.71} & \textbf{62.90} & \textbf{33.06} \\
 & \textit{w/o update} & 65.32 & 56.45 & 37.10 & 62.90 & 25.81 \\ 
\midrule

\multirow{4}{*}{\rotatebox[origin=c]{90}{\textbf{Calendar}}} 
 & Generic & 49.10 & 27.20 & 12.40 & 9.50 & 4.10 \\
 & CoT & 56.40 & 33.30 & \textbf{18.00} & 13.50 & 8.20 \\
 & React & 57.00 & 30.90 & 8.50 & 10.50 & 8.50 \\
 \cdashline{2-7}
 \noalign{\vskip 1.5pt} 
 & \textbf{IE as Cache} & \textbf{65.20} & \textbf{38.40} & 15.70 & \textbf{17.90} & \textbf{10.10} \\ 
\midrule

\multirow{4}{*}{\rotatebox[origin=c]{90}{\textbf{QMSUM}}} 
 & Generic & 32.03 & 32.92 & 28.76 & 31.82 & 27.12 \\
 & CoT & 31.70 & 31.53 & 27.54 & 29.30 & 28.44 \\
 & React & 29.38 & 29.59 & 24.27 & 27.61 & 28.64 \\
 \cdashline{2-7} 
 \noalign{\vskip 1.5pt} 
 & \textbf{IE as Cache} & \textbf{35.21} & \textbf{34.80} & \textbf{32.14} & \textbf{33.20} & \textbf{30.67} \\ 
\bottomrule
\end{tabular}%
}
\end{table}

%% file: tables/extraction.tex
\begin{table}[t]
\centering
\caption{Impact of extraction quality on final reasoning accuracy (EM) on the TACT dataset (GPT-4o). \textit{w/ gold schema} represents an oracle setting using ground-truth extraction schemas. Similarity is calculated via Sentence-T5-Large.}
\label{table:extraction_analysis}
\setlength{\tabcolsep}{4pt} 
\begin{tabular}{lcccc}
\toprule
\textbf{Method} & \textbf{ROUGE-1} & \textbf{ROUGE-L} & \textbf{Similarity} & \textbf{EM} \\
\midrule
IE as Tool & 27.68 & 21.65 & 84.90 & 61.29 \\
\textbf{IE as Cache} & 31.04 & 24.27 & 87.11 & 71.77 \\
\textit{w/ gold schema} & {50.35} & {39.55} & {91.97} & {73.39} \\
\bottomrule
\end{tabular}
\end{table}

%% file: sections/sec05-conclusion.tex
\section{Conclusion}

This paper repurposes Information Extraction (IE) from a static terminal objective to a dynamic \textit{cognitive infrastructure} for LLMs. We introduce the \textit{IE-as-Cache} framework, which manages extracted information as a read-write memory layer, effectively mitigating noise and context decay during multi-step reasoning. Our findings confirm that this mechanism significantly enhances performance across diverse benchmarks, improving both reasoning accuracy and efficiency. Ultimately, this study demonstrates that IE can be effectively utilized as a reusable cognitive resource, offering a promising direction for future research in the downstream applications of information extraction, particularly in noise-rich and unstructured contexts.

%% file: sections/secapp.tex
\appendices
\section{Broader Connections and Potential Downstream Implications of IE-as-Cache}

The main paper studies IE-as-Cache in text-centric agentic reasoning. More broadly, however, the core intuition of our framework---namely, that downstream intelligence benefits from compact, task-conditioned, and updateable intermediate structure---resonates with a wider body of recent work across retrieval, ranking, personalization, recommendation, planning, and alignment. We do not view these studies as direct baselines; rather, they help contextualize the broader applicability of our perspective.

\subsection{Evidence Selection, Retrieval, and Ranking as Controlled Intermediate Computation}

A first relevant direction concerns systems in which the quality of downstream generation depends critically on how evidence is selected, calibrated, and surfaced. This includes calibration for generative listwise reranking~\cite{lv2026learningemptinessdebiasinglistwise}, unified retrieval-and-ranking generation~\cite{zhang2025killingbirdsstoneunifying}, efficient personalized reranking with semi-autoregressive generation~\cite{cheng2026efficientpersonalizedrerankingsemiautoregressive}, and long-text generation pipelines that explicitly couple planning with information discovery~\cite{gu2025rapidefficientretrievalaugmentedlong}. Relatedly, recent efforts on evaluating interleaved RAG generation~\cite{zhang2025ragigbenchinnovativeevaluationragbased} and mitigating catastrophic forgetting in retrieval-augmented generation~\cite{huang2025selfaugmitigatingcatastrophicforgetting} further suggest that robust downstream reasoning depends not only on raw model capability, but also on how intermediate evidence states are organized, refreshed, and calibrated. From this angle, our IE-as-Cache framework can be seen as a complementary instance of controlled intermediate computation, where extracted structure serves as an explicit working memory rather than a static preprocessing artifact.

\subsection{Agentic Reasoning, Personalization, and Modular Control}

A second line of connection comes from recent work emphasizing modular control and user-conditioned reasoning in agentic systems. At the systems level, user-centric agent paradigms argue for placing control, memory, and private context closer to the user rather than treating models as monolithic platform services~\cite{zhang2026paradigmusercentricagentplatformcentric}. At the inference level, collaborative and personalized decoding frameworks such as CoSteer~\cite{lv2026costeercollaborativedecodingtimepersonalization} and SpecSteer~\cite{lv2026specsteersynergizinglocalcontext} show the value of decomposing generation into coordinated components with different roles. In interactive decision settings, thought-augmented planning agents~\cite{yu2025thoughtaugmentedplanningllmpoweredinteractive} similarly demonstrate the benefit of explicit planning signals over end-to-end opaque generation. These studies differ from our IE-centered formulation, but collectively reinforce a shared design principle: effective downstream intelligence often emerges from structured, updateable interfaces between memory, control, and reasoning rather than from a single undifferentiated context block.

\subsection{Potential Implications for Foundation Recommenders and Sequential Decision Systems}

Although our experiments focus on text reasoning, the underlying idea of maintaining a compact and semantically salient state may also be relevant to foundation recommender and sequential decision systems. Recent surveys and overviews point to the growing convergence between generation, retrieval, planning, and recommendation~\cite{wang2025generativelargerecommendationmodels,zhourui}. Within this landscape, user retrieval integration~\cite{shen2024exploringuserretrievalintegration}, adaptive cross-domain representation enhancement~\cite{zhang2024unifiedframeworkadaptiverepresentation}, large-scale recommendation architectures~\cite{ye2025fuxialphascalingrecommendationmodel,ye2026fuxilinearunleashingpowerlinear}, and scaling-law based analyses of sequential models~\cite{shen2025optimizingsequentialrecommendationmodels} all highlight the importance of managing high-dimensional user and item information through structured intermediate representations. Recent studies on reasoning behavior in foundation recommenders~\cite{zhang2026thinkinghurtsdiagnosingrectifying}, recursive self-improvement~\cite{zhang2026recommendersystemsteachthemselves}, and feature-generation paradigms~\cite{yin2025featureinteractionfeaturegeneration} further suggest that stronger models still benefit from better-organized internal or external states.

This broader pattern is also visible in industrial recommendation and prediction systems, where explicit architectural bias and compact representations remain crucial for efficiency and robustness, including dynamic low-order-aware fusion for CTR prediction~\cite{Wang_2025}, de-correlated expert networks~\cite{wang2025enhancingctrpredictiondecorrelated}, and embedding compression frameworks~\cite{wang2025universalframeworkcompressingembeddings}. While these works arise from domains beyond information extraction, they support the more general claim that downstream decision quality often depends on how much useful structure can be preserved in a compact, controllable, and updateable form.

\subsection{Connections to Alignment and Data Utility}

Finally, the broader theme of selectively preserving decision-relevant information also appears in recent work on alignment and data curricula. For example, SPARD~\cite{zhi2026spardselfpacedcurriculumrl} studies how reward dynamics and data utility can be jointly leveraged to improve RL alignment. Although this is a training-time perspective rather than an inference-time memory mechanism, it resonates with the same high-level intuition behind IE-as-Cache: intelligent systems benefit when the information kept active is filtered according to utility rather than naively accumulated.

This broader discussion does not imply that the above methods are direct counterparts of IE-as-Cache. Instead, it highlights a shared research trend across multiple domains: future intelligent systems may increasingly rely on explicit intermediate structure, adaptive evidence maintenance, and controllable memory-like abstractions to support robust reasoning and decision making.